\documentclass[10pt,twocolumn,letterpaper]{article}
\pdfoutput=1
\usepackage{cvpr}

\usepackage{times} 
\usepackage{epsfig}
\usepackage{url}  
\usepackage{lipsum}
\usepackage{amsmath, amsthm, amssymb, mathtools}
\usepackage{graphicx}  
\usepackage{multirow}
\usepackage{enumitem}
\setlist[itemize]{noitemsep, topsep=0pt}
\setlist[enumerate]{noitemsep, topsep=0pt}

\usepackage[colorlinks]{hyperref}

\cvprfinalcopy 

\def\cvprPaperID{3841} 

\ifcvprfinal\fi
\begin{document}

\title{TraPHic: Trajectory Prediction in Dense and Heterogeneous Traffic Using Weighted Interactions}
\author{Rohan Chandra\\
University of Maryland\\
College Park\\
\and
Uttaran Bhattacharya\\
University of Maryland\\
College Park\\
\and
Aniket Bera\\
University of North Carolina\\
Chapel Hill\\
\and
Dinesh Manocha\\
University of Maryland\\
College Park\\
Video, Code, and Dataset: \url{https://gamma.umd.edu/traphic}
}

\maketitle

\begin{abstract}
We present a new algorithm for predicting the near-term trajectories of road agents in dense traffic videos. Our approach is designed for heterogeneous traffic, where the road agents may correspond to buses, cars, scooters, bi-cycles, or pedestrians. We model the interactions between different road agents using a novel LSTM-CNN hybrid network for trajectory prediction. In particular, we take into account heterogeneous interactions that implicitly account for the varying shapes, dynamics, and behaviors of different road agents. In addition, we model horizon-based interactions which are used to implicitly model the driving behavior of each road agent. We evaluate the performance of our prediction algorithm, TraPHic, on the standard datasets and also introduce a new dense, heterogeneous traffic dataset corresponding to urban Asian videos and agent trajectories. We outperform state-of-the-art methods on dense traffic datasets by 30\%. Code for our implementation can be found on our project webpage.
\end{abstract}
\vspace{-15pt}

\section{Introduction}\label{sec:intro}

The increasing availability of cameras and computer vision techniques has made it possible to track traffic road agents in realtime. These road agents may correspond to vehicles such as cars, buses, or scooters as well as pedestrians, bicycles, or animals. The trajectories of road agents extracted from a video can be used to model traffic patterns, driver behaviors, that are useful for autonomous driving. In addition to tracking, it is also important to predict the future trajectory of each road agent in realtime. The predicted trajectories are useful for performing safe autonomous navigation, traffic forecasting, vehicle routing, and congestion management~\cite{schreier2014bayesian,DBLP:journals/corr/abs-1801-06523}.
 

\begin{figure}[t]
    \centering
    \includegraphics[width=\columnwidth]{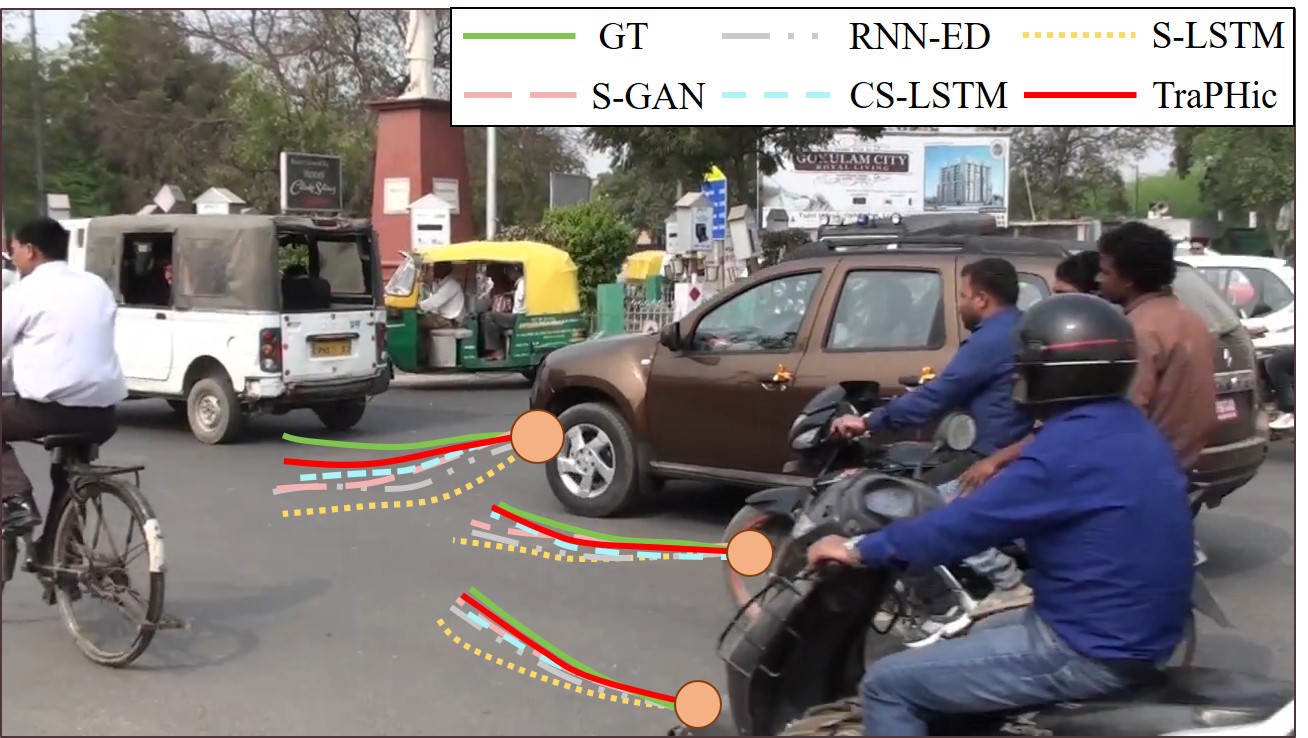}
    \caption{\textbf{Trajectory Prediction:} in dense heterogeneous traffic conditions. The scene consists of cars, scooters, motorcycles, three-wheelers, and bicycles in close proximity. Our algorithm (TraPHic) can predict the trajectory (red) of each road-agent close to the ground truth (green) and is better than other prior algorithms (shown in other colors).}
    \label{plc}
    \vspace{-15pt}
\end{figure}

In this paper, we deal with dense traffic composed of heterogeneous road agents. The heterogeneity corresponds to the interactions between different types of road agents such as cars, buses, pedestrians, two-wheelers (scooters and motorcycles), three-wheelers (rickshaws), animals, etc. These agents have different shapes, dynamic constraints, and behaviors. The traffic density corresponds to the number of distinct road agents captured in a single frame of the video or the number of agents per unit length (\textit{e.g.}, a kilometer) of the roadway. High density traffic is described as traffic with more than 100 road agents per Km. Finally, an interaction corresponds to how two road agents in close proximity affect each other's movement or avoid collisions.

There is considerable work on trajectory prediction for moving agents~\cite{social-lstm,social-gan,vermula2018social,nachiket,lee2017desire, tp,nachiket,lee2017desire}.
Most of these algorithms have been developed for scenarios with single type of agents (\textit{a.k.a.} homogeneous agents), which may correspond to human pedestrians in a crowd or cars driving on a highway. Furthermore, many prior methods have been evaluated on traffic videos corresponding to relatively sparse scenarios with only a few heterogeneous interactions, such as the NGSIM \cite{ngsim} and KITTI \cite{kitti} datasets. In these cases, the interaction between agents can be modeled using well-known models based on social forces \cite{earlysocial}, velocity obstacles \cite{van2011reciprocal}, or LTA \cite{eth}.

Prior prediction algorithms do not work well on dense, heterogeneous traffic scenarios because they do not model the interactions accurately. For example, the dynamics of a bus-pedestrian interaction differs significantly from a pedestrian-pedestrian or a car-pedestrian interaction due to the differences in shape, size, maneuverability, and velocities. The differences in the dynamic characteristics of road agents affect their trajectories and how they navigate around each other in dense traffic situations~\cite{autorvo}.
Moreover, prior learning-based prediction algorithms typically model the interactions uniformly for all other road agents in its neighborhood and the resulting model assigns equal weight to each interaction. This method works well for homogeneous traffic. However, it does not work well for dense heterogeneous traffic, and we need methods to assign different weights to different pairwise interactions.  

\textbf{Main Contributions:} We present a novel traffic prediction algorithm, TraPHic, for predicting the trajectories of road agents in realtime. The input to our algorithm is the trajectory history of each road agent as observed over a short time-span ($2$-$4$ seconds), and the output is the predicted trajectory over a short span ($3$-$5$ seconds). In order to develop a general approach to handle dense traffic scenarios, our approach models two kinds of weighted interactions, horizon-based and heterogeneous-based.
\begin{enumerate}
\item \textit{Heterogeneous-Based:} We implicitly take into account varying sizes, aspect ratios, driver behaviors, and dynamics of road agents. Our formulation accounts for several dynamic constraints such as average velocity, turning radius, spatial distance from neighbors, and local density. We embed these functions into our state-space formulation and use them as inputs to our network to perform learning.
 
\item \textit{Horizon-Based:} We use a semi-elliptical region (horizon) based on a pre-defined radius in front of each road agent. We prioritize the interactions in which the road agents are within the horizon using a Horizon Map. Our approach learns a weighting mechanism using a non-linear formulation, and uses that to assign weights to each road agent in the horizon automatically.
\end{enumerate}

We formulate these interactions within an LSTM-CNN hybrid network that learns locally useful relationships between the heterogeneous road agents. Our approach is end-to-end and does not require explicit knowledge of an agent's behavior. Furthermore, we present a new traffic dataset (TRAF) comprising of dense and heterogeneous traffic. The dataset consists of the following road agents: cars, buses, trucks, rickshaws, pedestrians, scooters, motorcycles, carts, and animals and is collected in dense Asian cities. We also compare our approach with prior methods and highlight the accuracy benefits. Overall, TraPHiC offers the following benefits as a realtime prediction algorithm:
\begin{enumerate}
    \item TraPHIC outperforms prior methods on dense traffic datasets with $10$-$30$ road agents by $0.78$ meters on the root mean square error~(RMSE) metric, which is a $30\%$ improvement over prior methods.
    \item Our algorithm offers accuracy similar to prior methods on sparse or homogeneous datasets such as the NGSIM dataset~\cite{ngsim}.
\end{enumerate}
The rest of the paper is organized as follows. We give a brief overview of prior work in Section~\ref{sec:rw}. Section~\ref{sec:traphic} presents an overview of the weighted interactions. We present the overall learning algorithm in Section~\ref{sec:arch} and evaluate its performance on different datasets in Section~\ref{sec:results}.

\section{Related Work}\label{sec:rw}

In this section, we give a brief overview of some important classical prediction algorithms and recent techniques based on deep neural networks.

\subsection{Prediction Algorithms and Interactions}
\vspace{-5pt}
Trajectory prediction has been researched extensively. Approaches include the Bayesian formulation \cite{early1-bayesian}, the Monte Carlo simulation \cite{early5}, Hidden Markov Models (HMMs) \cite{early2-HMM}, and Kalman Filters \cite{early4-kalman}.

Methods that do not model road-agent interactions are regarded as sub-optimal or as less accurate than methods that  model the interactions between road agents in the scene~\cite{early3-unfreeze}.
Examples of methods that explicitly model road-agent interaction include techniques based on  social forces~\cite{earlysocial,early6}, velocity obstacles~\cite{van2011reciprocal}, LTA~\cite{eth}, etc. Many of these models were designed to account for interactions between pedestrians in a crowd (\textit{i.e.} homogeneous interactions) and improve the prediction accuracy~\cite{bera2016glmp}.
Techniques based on velocity obstacles have been extended using kinematic constraints to model the interactions between heterogeneous road agents~\cite{autorvo}. Our learning approach does not use any explicit pairwise motion model. Rather, we model the heterogeneous interactions between road agents implicitly.
\subsection{Deep-Learning Based Methods}
\vspace{-5pt}
Approaches based on deep neural networks use variants of Recurrent Neural Networks (RNNs) for sequence modeling. These have been extended to hybrid networks by combining RNNs with other deep learning architectures for motion prediction.

\noindent\textbf{RNN-Based Methods} 
RNNs are natural generalizations of feedforward neural networks to sequence~\cite{rnn1}. The benefits of RNNs for sequence modeling makes them a reasonable choice for traffic prediction. Since RNNs are incapable of modeling long-term sequences, many traffic trajectory prediction methods use long short-term memory networks (LSTMs) to model road-agent interactions. These include algorithms to predict trajectories in traffic scenarios with few heterogeneous interactions~\cite{nachiket,tp}. These techniques have also been used for trajectory prediction for pedestrians in a crowd~\cite{social-lstm, vermula2018social}.

\noindent\textbf{Hybrid Methods}
Deep-learning-based hybrid methods consist of networks that integrate two or more deep learning architectures. Some examples of deep learning architectures include CNNs, GANs, VAEs, and LSTMs. Each architecture has its own advantages and, for many tasks, the advantages of individual architectures can be combined. There is considerable work on the development of hybrid networks. Generative models have been successfully used for tasks such as super resolution~\cite{hybridsuper}, image-to-image translation~\cite{hybridit}, and image synthesis~\cite{hybriddraw}. However, their application in trajectory prediction has been limited because back-propagation during training is non-trivial. In spite of this, generative models such as VAEs and GANs have been used for trajectory prediction of pedestrians in a crowd~\cite{social-gan} and in sparse traffic \cite{lee2017desire}. Alternatively, Convolutional Neural Networks (CNNs or ConvNets) have also been successfully used in many computer vision applications like object recognition \cite{objrecogreview}. Recently, they have also been used for traffic trajectory prediction \cite{cnnpredict1,cnnpredict2}. In this paper, we present a new hybrid network that combines LSTMs with CNNs for traffic prediction. 
\vspace{-5pt}
\subsection{Traffic Datasets}
\vspace{-5pt}
There are several datasets corresponding to traffic scenarios. ApolloScape \cite{datasetapollo} is a large-scale
dataset of street views that contain scenes with higher complexities, 2D/3D annotations and pose information, lane markings and video frames. However, this dataset does not provide trajectory information. The NGSIM simulation dataset~\cite{ngsim} consists of trajectory data for road agents corresponding to cars and trucks, but the traffic scenes are limited to highways with fixed-lane traffic. KITTI~\cite{kitti} dataset has been used in different computer vision applications such as stereo, optical flow, 2D/3D object detection, and tracking. There are some pedestrian trajectory datasets like ETH~\cite{eth} and UCY~\cite{ucy}, but they are limited to pedestrians in a crowd. Our  new dataset, TRAF, corresponds to dense and heterogeneous traffic captured from Asian cities and includes 2D/3D trajectory information. 

\section{TraPHic: Trajectory Prediction in Heterogeneous Traffic}\label{sec:traphic}

\begin{figure}[t]

\centering
    \includegraphics[width=.75\columnwidth]{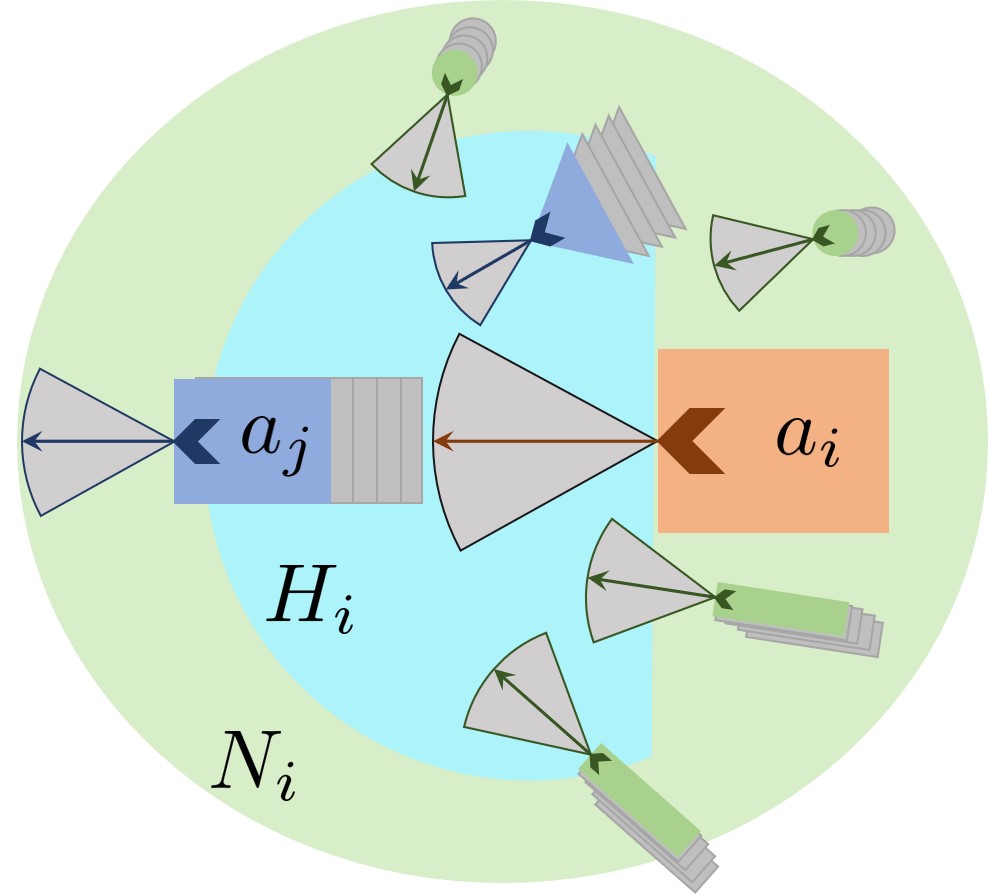} \label{interactions}
    
\caption{\textbf{Horizon and Heterogeneous Interactions: }We highlight various interactions for the red car. Horizon-based weighted interactions are in the blue region, containing a car and a rickshaw (both blue). The red car prioritizes the interaction with the blue car and the rickshaw (\textit{i.e.} avoids a collision) over interactions with other road-agents. Heterogeneous-Based weighted interactions are in the green region, containing pedestrians and motorcycles (all in green). We model these interactions as well to improve the prediction accuracy.} 
\label{horizon}
\end{figure}

In this section, we give an overview of our prediction algorithm that uses weighted interactions. Our approach is designed for dense and heterogeneous traffic scenarios and is based on two observations. The first observation is based on the idea that road agents in such dense traffic do not react to every road agent around them; rather, they selectively focus attention on key interactions in a semi-elliptical region in the field of view, which we call the ``horizon''. For example, consider a motorcyclist who suddenly moves in front of a car and the neighborhood of the car consists of other road agents such as three-wheelers and pedestrians (Figure~\ref{horizon}). The car must prioritize the motorcyclist interaction over the other interactions to avoid a collision.

The second observation stems from the heterogeneity of different road agents such as cars, buses, rickshaws, pedestrians, bicycles, animals, etc. in the neighborhood of an road agent (Figure~\ref{horizon}). For instance, the dynamic constraints of a bus-pedestrian interaction differs significantly from a pedestrian-pedestrian or even a car-pedestrian interaction due to the differences in road agent shapes, sizes, and maneuverability. To capture these heterogeneous road agent dynamics, we embed these properties into the state-space representation of the road agents and feed them into our hybrid network. We also implicitly model the behaviors of the road agents.
Behavior in our case the different driving and walking styles of different drivers and pedestrians. Some are more aggressive while others more conservative.
We model these behaviors as they directly influence the outcome of various interactions~\cite{aniketdriver}, thereby affecting the road agents' navigation.


\subsection{Problem Setup and Notation}

Given a set of $N$ road agents $\mathcal{A}=\{a_i\}_{i=1\dots N}$, trajectory history of each road agent $a_i$ over $t$ frames, denoted $\Psi_{i,t} := [(x_{i,1},y_{i,1}),\dots,(x_{i,t},y_{i,t})]^\top$, and the road agent's size $l_i$, we predict the spatial coordinates of that road agent for the next $\tau$ frames. In addition, we introduce a feature called traffic concentration $c$, motivated by traffic flow theory~\cite{iannini2016kinetic}. Traffic concentration, $c(x, y)$, at the location $(x, y)$ is defined as the number of road agents between $(x, y)$ and $(x, y)+(\delta x, \delta y)$ for some predefined $(\delta x, \delta y) > 0$. This metric is similar to traffic density, but the key difference is that traffic density is a macroscopic property of a traffic video, whereas traffic concentration is a mesoscopic property and is locally defined at a particular location. So we achieve a representation of traffic on several scales.

Finally, we define the state space of each road agent $a_i$ as
\begin{equation}
    \Omega_i := \begin{bmatrix}\Psi_{i, t} & \Delta\Psi_{i, t} & c_i & l_i\end{bmatrix}^\top
\end{equation}
where $\Delta$ is a derivative operator that is used to compute the velocity of the road agent, and $c_i:=[c(x_{i,1},y_{i,1}),\dots,c(x_{i,t},y_{i,t})]^\top$.

 
\noindent \textbf{2D Image Space to 3D World Coordinate Space: } We compute camera parameters from given videos using standard techniques~\cite{cameracalib1, cameracalib2}, and use the parameters to estimate the camera homography matrices. The homography matrices are subsequently used to convert the location of road agents in 2D pixels to 3D world coordinates w.r.t. a predetermined frame of reference, similar to approaches in \cite{social-gan,social-lstm}. All state-space representations are subsequently converted to the 3D world space.

\noindent \textbf{Horizon and Neighborhood Agents: } Prior trajectory prediction methods have collected neighborhood information using lanes and rectangular grids~\cite{nachiket}. Our approach is more generalized in that we pre-process the trajectory data by assuming a lack of lane information. This assumption is especially true in practice in dense and heterogeneous traffic conditions. We formulate a road agent $a_i$'s neighborhood, $N_i$, using an elliptical region and selecting a fixed number of closest road agents using the nearest-neighbor search algorithm in that region. Similarly, we define the horizon of that agent, $H_i$, by selecting a smaller threshold in the nearest-neighbor search algorithm, and in a semi-elliptical region in front of $a_i$.
\begin{figure*}
    \centering
    \includegraphics[width=\textwidth]{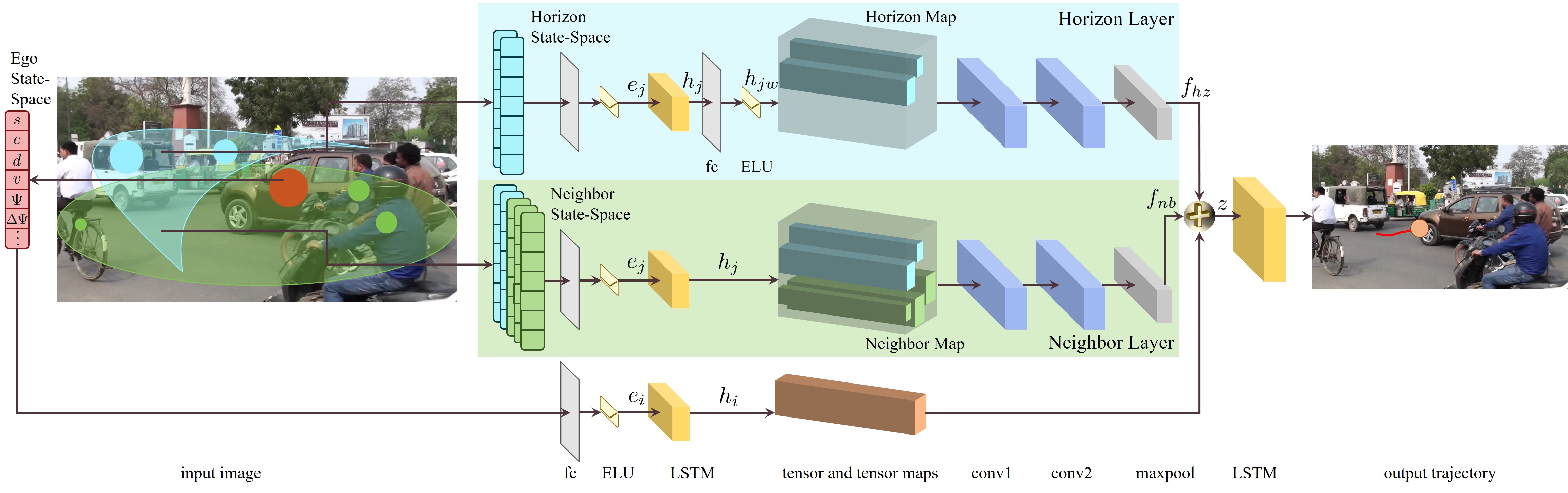}
    \caption{\textbf{TraPHic Network Architecture:} The ego agent is marked by the red dot. The green elliptical region around it is its neighborhood and the cyan semi-elliptical region in front of it is its horizon. We generate input embeddings for all agents based on trajectory information and heterogeneous dynamic constraints such as agent shape, velocity, and traffic concentration at the agent's spatial coordinates, and other parameters. These embeddings are passed through LSTMs and eventually used to construct the horizon map, the neighbor map and the ego agent's own tensor map. The horizon and neighbor maps are passed through separate ConvNets and then concatenated together with the ego agent tensor to produce latent representations. Finally, these latent representations are passed through an LSTM to generate a trajectory prediction for the ego agent.}
    \label{main}
    \vspace{-10pt}
\end{figure*}


\section{Hybrid Architecture for Traffic Prediction}\label{sec:arch}
\vspace{-5pt}
In this section, we present our novel network architecture for performing trajectory prediction in dense and heterogeneous environments. In the context of heterogeneous traffic, the goal is to predict trajectories,  \textit{i.e.} temporal sequences of spatial coordinates of a road agent. Temporal sequence prediction requires models that can capture temporal dependencies in data, such as LSTMs~\cite{graves}. However, LSTMs cannot learn dependencies or relationships of various heterogeneous road agents because the parameters of each individual LSTM are independent of one another. In this regard, ConvNets have been used in computer vision applications with greater success because they can learn locally dependent features from images. Thus, in order to leverage the benefits of both, we combine ConvNets with LSTMs to learn locally useful relationships, both in space and in time, between the heterogeneous road agents. We now describe our model to predict the trajectory for each road agent $a_i$. A visualization of the model is shown in Figure~\ref{main}.

We start by computing $H_i$ and $N_i$ for the agent $a_i$. Next, we identify all road agents $a_j \in N_i \cup H_i$. Each $a_j$ has an input state-space $\Omega_j$ that is used to create the embeddings $e_j$, using
\begin{equation}
    e_j = \phi(W_l\Omega_i + b_l)
\end{equation}
where $W_l$ and $b_l$ are conventional symbols denoting the weight matrix and bias vector respectively, of the layer $l$ in the network, and $\phi$ is the non-linear activation on each node.

Our network consists of three layers. The horizon layer~(top cyan layer in Figure~\ref{main}) takes in the embedding of each road agent in $H_i$, and the neighbor layer~(middle green layer in Figure~\ref{main}) takes in the embedding of each road agent in $N_i$. The input embeddings in both these layers are passed through fully connected layers with ELU non-linearities~\cite{elu}, and then fed into single-layered LSTMs (yellow blocks in Figure \ref{main}). The outputs of the LSTMs in the two layers are hidden state vectors, $h_j(t)$, that are computed using
\begin{equation}
    h_j(t) = \textrm{LSTM}(e_j, W_l, b_l, h_j^{t-1})
\end{equation}
\noindent where $h_j^{t-1}$ refers to the corresponding road agent's hidden state vector from the previous time-step $t-1$. The hidden state vector of a road agent is a latent representation that contains temporally useful information. In the remainder of the text, we drop the parameter $t$ for the sake of simplicity, \textit{i.e.}, $h_j$ is understood to mean $h_j(t)$ for any $j$.

The hidden vectors in the horizon layer are passed through an additional fully connected layer with ELU non-linearities~\cite{elu}. We denote the output of the fully connected layer as $h_{jw}$. All the $h_{jw}$'s in the horizon layer are then pooled together in a ``horizon map''. The hidden vectors in the neighbor layer are directly pooled together in a ``neighbor map''. These maps are further elaborated in Section~\ref{sub:weighted_interactions}. Both these maps are then passed through separate ConvNets in the two layers. The ConvNets in both the layers are comprised of two convolution operations followed by a max-pool operation. We denote the output feature vector from the ConvNet in the horizon layer as $f_\textrm{hz}$, and that from the ConvNet in the neighbor layer as $f_\textrm{nb}$.

Finally, the bottom-most layer corresponds to the ego agent $a_i$. Its input embedding, $e_i$, passes sequentially through a fully connected with ELU non-linearities~\cite{elu}, and a single-layered LSTM to compute its hidden vector, $h_i$. The feature vectors from the horizon and neighbor layers, $f_\textrm{hz}$ and $f_\textrm{nb}$, are concatenated with $h_i$ to generate a final vector encoding
\begin{equation}
    z := \textrm{concat}(h_i, f_\textrm{hz}, f_\textrm{nb})
\end{equation}
\noindent Finally, the concatenated encoding $z$ passes through an LSTM to compute the prediction for the next $\tau$ seconds.

\subsection{Weighted Interactions}\label{sub:weighted_interactions}
Our model is trained to learn weighted interactions in both the horizon and neighborhood layers. Specifically, it learns to assign appropriate weights to various pairwise interactions based on the shape, dynamic constraints and behaviors of the involved agents. The \textit{horizon-based weighted interactions} takes into account the agents in the horizon of the ego agent, and learns the ``horizon map'' $\mathcal{H}_i$, given as
\begin{equation}
    \mathcal{H}_i = \{ h_{jw} | a_j \in H_i \}
\end{equation}
\noindent Similarly, the neighbor or \textit{heterogeneous-based weighted interactions} accounts for all the agents in the neighborhood of the ego agent, and learns the ``neighbor map'' $\mathcal{N}_i$, given as
\begin{equation}
    \mathcal{N}_i = \{ h_j | a_j \in N_i \}
\end{equation}
\noindent During training, back-propagation optimizes the weights corresponding to these maps by minimizing the loss between predicted output and ground truth labels. Our formulation results in higher weights for prioritized interactions (larger tensors in Horizon Map or blue vehicles in Figure~\ref{horizon}) and lower weights for less relevant interactions (smaller tensors in Neighbor Map or green vehicles in Figure~\ref{horizon}).

\subsection{Implicit Constraints}
\noindent \textbf{Turning Radius:}
In addition to constraints such as position, velocity and shape, constraints such as the turning radius of a road agent also affects its maneuverability, especially as it interacts with other road agents within some distance. For example, a car (a non-holonomic agent) cannot alter its orientation in a short time frame to avoid collisions, whereas a bicycle or a pedestrian can.

However, the turning radius of a road agent can be determined by the dimensions of the road agent, \textit{i.e.}, its length and width. Since we include these parameters into our state-space representation, we implicitly take into consideration each agent's turning radius constraints as well.

\noindent \textbf{Driver Behavior:}
As stated in \cite{aniketdriver}, velocity and acceleration (both relative and average ) are clear indicators of driver aggressiveness. For instance, a road agent with a relative velocity (and/or acceleration) much higher than the average velocity (and/or acceleration) of all road agents in a given traffic scenario would be deemed as aggressive. Moreover, given the traffic concentrations at two consecutive spatial coordinates, $c(x, y)$ and $c(x+\delta x, y+\delta y)$, where $c(x, y) >> c(x+\delta x, y+\delta y)$, aggressive drivers move in a ``greedy'' fashion in an attempt to occupy the empty spots in the subsequent spatial locations. For each road agent, we compute its concentration with respect to its neighborhood and add this value to its input state-space.

Finally, the relative distance of a road agent from its neighbors is another factor pertaining to how conservative or aggressive a driver is. More conservative drivers tend to maintain a healthy distance while aggressive drivers tend to tail-gate. Hence, we compute the spatial distance of each road agent in the neighborhood and encode this in its state-space representation.

\subsection{Overall Trajectory Prediction}
Our algorithm follows a well-known scheme for prediction~\cite{social-lstm}. We assume that the position of the road agent in the next frame follows a bi-variate Gaussian distribution with parameters $\mu_i^t, \sigma_i^t = [(\mu_x, \mu_y)_i^t, ((\sigma_x, \sigma_y)_i^t)]$, and correlation coefficient $\rho_i^t$. The spatial coordinates $(x_i^t,y_i^t)$ are thus drawn from $\mathcal{N}(\mu_i^t, \sigma_i^t, \rho_i^t)$. We train the model by minimizing the negative log-likelihood loss function for the $i^{\textrm{th}}$ road agent trajectory,
\begin{equation}
    L_i = - \Sigma_{t+1}^\tau \log (\mathbf{P}((x_i^t,y_i^t)|(\mu_i^t, \sigma_i^t, \rho_i^t))).
\end{equation}

\noindent We jointly back-propagate through all three layers of our network, optimizing the weights for the linear blocks, ConvNets, LSTMs, and Horizon and Neighbor Maps. The optimized parameters learned for the Linear-ELU block in the horizon layer indicates the priority for the interaction in the horizon of an road agent $a_i$.


\section{Experimental Evaluation}\label{sec:results}
\begin{table}
    \centering
    \resizebox{\columnwidth}{!}{
    \begin{tabular}{|c|c|c|c|c|c|}
        \hline
        Dataset & \multicolumn{5}{c|}{Method} \\
        \cline{2-6}
        & RNN-ED & S-LSTM & S-GAN & CS-LSTM  &\textbf{TraPHic}  \\
        \hline
        NGSIM & 6.86/10.02 & 5.73/9.58 & \textbf{5.16/9.42} & 7.25/10.05 & 5.63/9.91  \\
        \hline
       Beijing & 2.24/8.25 & 6.70/8.08 & 4.02/7.30 & 2.44/8.63  & \textbf{2.16/6.99} \\
        \hline
    \end{tabular}
    }
    \caption{\textbf{Evaluation} on sparse or homogeneous traffic datasets: The first number is the average RMSE error (ADE) and the second number is final RMSE error (FDE) after $5$ seconds (in meters). NGSIM is a standard sparse traffic dataset with few heterogeneous interactions. The Beijing dataset is dense but with relatively low heterogeneity. Lower value is better and bold value represents the most accurate result.}
    
    \label{standard}
    \vspace{-10pt}
\end{table}


\begin{table*}
    \centering
        \resizebox{\textwidth}{!}{
    \begin{tabular}{|c|c|c|c|c|c|c|c|c|c|c|}
        \hline
        \multicolumn{11}{|c|}{Methods Evaluated on TRAF}\\
        \cline{1-11}
         RNN-ED & \multicolumn{2}{c|}{S-LSTM} & \multicolumn{2}{c|}{S-GAN} & \multicolumn{2}{c|}{CS-LSTM}  &\multicolumn{4}{c|}{\textbf{TraPHic}} \\
        \cline{2-11}
         & Original & Learned & Original & Learned & Original & Learned & B & $H_e$ & $H_o$ & Combined \\
        \hline
         3.24/5.16 & 6.43/6.84 & 3.01/4.89 & 2.89/4.56 & 2.76/4.79 & 2.34/8.01 & 1.15/3.35 & 2.73/7.21 & 2.33/5.75 & 1.22/3.01 & \textbf{0.78/2.44} \\
        \hline
    \end{tabular}
    }
        \caption{\textbf{Evaluation} on our new, highly dense and heterogeneous TRAF dataset. The first number is the average RMSE error (ADE) and the second number is final RMSE error (FDE) after $5$ seconds (in meters). The original setting for a method indicates that it was tested with default settings. The learned setting indicates that it was trained on our dataset for fair comparison. We present variations of our approach with each weighted interaction and demonstrate the contribution of the method. Lower is better and bold is best result.}
        \label{traf}
        \vspace{-8pt}
\end{table*}


\begin{table*}[!htb]
    \centering
    \resizebox{\textwidth}{!}{
  \begin{tabular}{|l|c|c|c|c|c|c|c|c|c|c|c|c|c|}
  \hline
    \multirow{2}{*}{Dataset} & \# Frames & \multicolumn{9}{c|}{Agents} & Visibility & Density & \#Diff \\
    \cline{3-11}
    & $(\times 10^3)$ & Ped & Bicycle & Car & Bike & Scooter & Bus & Truck & Rick & Total & (Km) & $(\times 10^{3})$ & Agents \\
    \hline
    NGSIM & 10.2 & 0 & 0 & 981.4 & 3.9 & 0 & 0 & 28.2 & 0 & 1013.5 & 0.548 & 1.85 & 3\\
    \hline
    Beijing & 93 & 1.6 & 1.9 & \multicolumn{6}{c|}{12.9} & 16.4 & 0.005 & 3.28 & 3\\
    \hline
    \textbf{TRAF} & 12.4 & 4.9 & 1.5  &3.6 & 1.43 & 5 & 0.15 & 0.2 & 3.1 & 19.88 & 0.005 & 3.97 & 8 \\
    \hline
  \end{tabular}
  }
    \caption{\textbf{Comparison} of our new TRAF dataset with various traffic datasets in terms of heterogeneity and density of traffic agents. Heterogeneity is described in terms of the number of different agents that appear in the overall dataset. Density is the total number of traffic agents per Km in the dataset. The value for each agent type under ``Agents'' corresponds to the average number of instances of that agent per frame of the dataset. It is computed by taking all the instances of that agent and dividing by the total number of frames. Visibility is a ballpark estimate of the length of road in meters that is visible from the camera. NGSIM data were collected using tower-mounted cameras (bird's eye view), whereas both Beijing and TRAF data presented here were collected with car-mounted cameras (frontal view).}
    \label{dataset}
\end{table*}

We describe our new dataset in Section~\ref{sub: TRAF}. In Section~\ref{sub: imple}, we list all implementation details used in our training process. Next, we list the evaluation metrics and methods that we compare with, in Section~\ref{sub:eval methods}. Finally, we present the evaluation results in Section~\ref{sub: results}.

\subsection{TRAF Dataset: Dense \& Heterogeneous Urban Traffic}
\label{sub: TRAF}
We present a new dataset, currently comprising of $50$ videos of dense and heterogeneous traffic. The dataset consists of the following road agent categories: car, bus, truck, rickshaw, pedestrian, scooter, motorcycle, and other road agents such as carts and  animals. Overall, the dataset contains approximately $13$ motorized vehicles, $5$ pedestrians and $2$ bicycles per frame. Annotations were performed following a strict protocol and each annotated video file consists of spatial coordinates, an agent ID, and an agent type. The dataset is categorized according to camera viewpoint (front-facing/top-view), motion (moving/static), time of day (day/evening/night), and difficulty level (sparse/moderate/heavy/challenge). All the videos have a resolution of $1280\times 720$. We present a comparison of our dataset with standard traffic datasets in Table \ref{dataset}. The dataset is available at \url{https://gamma.umd.edu/traphic/dataset}.

\subsection{Implementation Details}
\label{sub: imple}
We use single-layer LSTMs as our encoders and decoders with hidden state dimensions of $64$ and $128$, respectively. Each ConvNet is implemented using two convolutional operations each followed by an ELU non-linearity~\cite{elu} and then max-pooling. We train the network for $16$ epochs using the Adam optimizer~\cite{kingma2014adam} with a batch size of $128$ and learning rate of $0.001$. We use a radius of $2$ meters to define the neighborhood and a minor axis length of $1.5$ meters to define the horizon, respectively. Our approach uses $3$ seconds of history and predicts spatial coordinates of the road agent for up to $5$ seconds ($4$ seconds for KITTI dataset). We do not down-sample on the NGSIM dataset due to its sparsity. However, we use a down-sampling factor of $2$ on the Beijing and TRAF datasets due to their high density. Our network is implemented in Pytorch using a single TiTan Xp GPU. Our network does not use batch norm or dropout as they can decrease accuracy. We include the experimental details involving batch norm and dropout in the appendix due to space limitations.

\subsection{Evaluation Metrics and Comparison Methods}
\label{sub:eval methods}
We use the following commonly used metrics~\cite{social-lstm,social-gan,nachiket} to measure the performances of the algorithms used for predicting the trajectories of the road agents.
\begin{enumerate}
    \item Average displacement error (ADE): The root mean square error
(RMSE) of all the predicted positions and real positions
during the prediction time.
\item Final displacement error (FDE): The RMSE distance between the final predicted positions at the end of the predicted trajectory and the corresponding true location.
\end{enumerate}
We compare our approach with the following methods.
\begin{itemize}
    \item RNN-ED (Seq2Seq): An RNN encoder-decoder model, which
is widely used in motion and trajectory prediction for vehicles~\cite{Britz:2017}.
    \item Social-LSTM (S-LSTM): An LSTM-based network with social
pooling of hidden states to predict pedestrian trajectories in crowds~\cite{social-lstm}. 
    \item Social-GAN (S-GAN): An LSTM-GAN hybrid network to predict trajectories for large human crowds~\cite{social-gan}.
    \item Convolutional-Social-LSTM (CS-LSTM): A variant of S-LSTM adding convolutions to the network in~\cite{social-lstm} in order to predict trajectories in sparse highway traffic~\cite{nachiket}.
\end{itemize}
We also perform ablation studies with the following four versions of our approach.
\begin{itemize}
    \item TraPHic-B: A base version of our approach without using any weighted interactions.
    \item TraPHic-$H_o$: A version of our approach without using \textit{Heterogeneous}-Based Weighted interactions, \textit{i.e.}, we do not take into account driver behavior and information such as shape, relative velocity, and concentration.
    \item TraPHic-$H_e$: A version of our approach without using \textit{Horizon}-Based Weighted interactions. In this case, we do not explicitly model the horizon, but account for heterogeneous interactions.
    \item TraPHic: Our main algorithm using both Heterogeneous-Based and Horizon-Based Weighted interactions. We explicitly model the horizon and implicitly account for dynamic constraints and driver behavior.
\end{itemize}
\begin{figure}[b]
    \centering
    \includegraphics[width=\columnwidth]{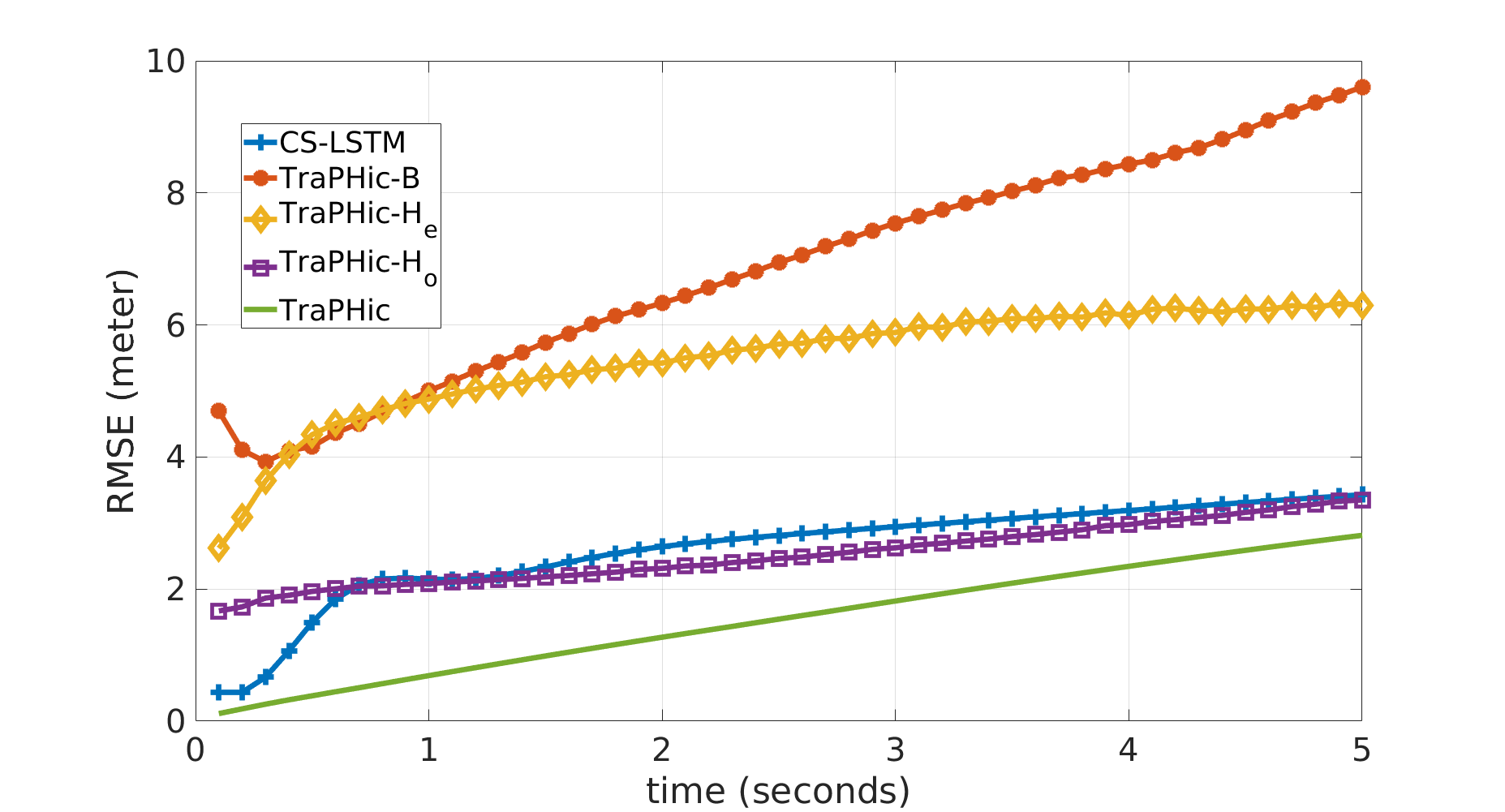}
    \caption{\textbf{RMSE Curve Plot: }We compare the accuracy of four variants of our algorithm with CS-LSTM and each other based on RMSE values on the TRAF dataset. On the average, using TraPHic-$H_e$ reduces RMSE by $15\%$ relative to TraPHic-B, and using TraPHic-$H_o$ reduces RMSE by $55\%$ relative to TraPHic-B. TraPHic, the combination of TraPhic-$H_e$ and TraPhic-$H_o$, reduces RMSE by $36\%$ relative to TraPHic-$H_o$, $66\%$ relative to TraPHic-$H_e$, and $71\%$ relative to TraPHic-B. Relative to CS-LSTM, TraPHic reduces RMSE by $30\%$.}
    \label{rmse}
    \vspace{-10pt}
\end{figure}

\begin{figure*}
    \centering
    \includegraphics[width=\textwidth]{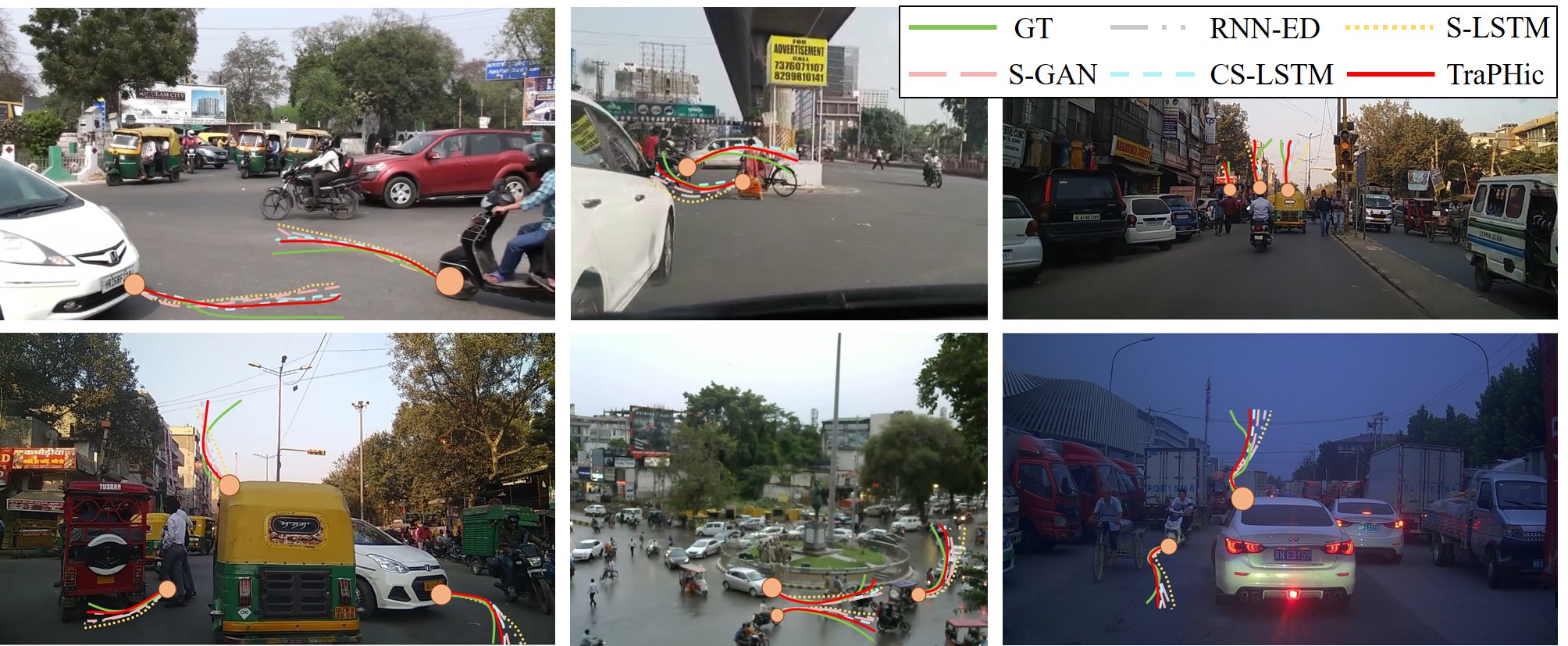}
    \caption{\textbf{Trajectory Prediction Results: }We highlight the performance of various trajectory prediction methods on our TRAF dataset with different types of road-agents. We showcase six scenarios with different density, heterogeneity, camera position (fixed or moving), time of the day, and weather conditions. We highlight the predicted trajectories (over $5$ seconds) of some of the road-agents in each scenario to avoid clutter. The ground truth (GT) trajectory is drawn as a solid green line, and our (TraPHic) prediction results are shown using a solid red line. The prediction results of other methods (RNN-ED, S-LSTM, S-GAN, CS-LSTM) are drawn with different dashed lines. TraPHic predictions are closest to  GT in all the scenarios. We observe up to $30\%$ improvement in accuracy over prior methods over this dense, heterogeneous traffic.}
    \label{visual_res}
    \vspace{-10pt}
\end{figure*}

\subsection{Results on Traffic Datasets}
\label{sub: results}
In order to provide a comprehensive evaluation, we compare our method with state-of-the-art methods on several datasets. Table~\ref{standard} shows the results on the standard NGSIM dataset and an additional dataset containing heterogeneous traffic of moderate density. We present results on our new TRAF dataset in Table~\ref{traf}.

TraPHic outperforms all prior methods we compared with on our TRAF dataset. For a fairer comparison, we trained these methods on our dataset before testing them on the dataset. However, the prior methods did not generalize well to dense and heterogeneous traffic videos. One possible explanation for this is that S-LSTM and S-GAN were designed to predict trajectories of humans in top-down crowd videos whereas the TRAF dataset consists of front-view heterogeneous traffic videos with high density. CS-LSTM uses lane information in its model and weight all agent interactions equally. Since the traffic in our dataset does not include the concept of lane-driving, we used the version of CS-LSTM that does not include lane information for a fairer comparison. However, it still led to a poor performance since CS-LSTM does not account for heterogeneous-based interactions. On the other hand, TraPHic considers both heterogeneous-based and horizon-based interactions, and thus produces superior performance on our dense and heterogeneous dataset.

We visualize the performance of the various trajectory prediction methods on our TRAF dataset Figure~\ref{visual_res}. Compared to the prior methods, TraPHic produces the least deviation from the ground truth trajectory in all the scenarios. Due to the significantly high density and heterogeneity in these videos, coupled with the unpredictable nature of the involved agents, all the predictions deviate from the ground truth in the long term (after $5$ seconds).

We demonstrate that our approach is comparable to prior methods on sparse datasets such as the NGSIM dataset. We do not outperform the current sate-of-the-art in such datasets, since our algorithm tries to account for heterogeneous agents and weighted interactions even when interactions are sparse and mostly homogeneous. Nevertheless, we are at par with the state-of-the-art performance. Lastly, we note that our RMSE value on the NGSIM dataset is quite high, which we attribute to the fact that we used a much higher (2X) sampling rate for averaging than prior methods.

Finally, we perform an ablation study to highlight the contribution of our weighted interaction formulation. We compare the four versions of TraPHic as stated in Section~\ref{sub:eval methods}. We find that the Horizon-based formulation contributes more significantly to higher accuracy. TraPHic-$H_e$ reduces ADE by $15\%$ and FDE by $20\%$ over TraPHic-B, whereas TraPHic-$H_o$ reduces ADE by $55\%$ and FDE by $58\%$ over TraPHic-B. Incorporating both formulations results in the highest accuracy, reducing the ADE by $71\%$ and the FDE by $66\%$ over TraPHic-B.


\vspace{-7pt}
\section{Conclusion, Limitations, and Future Work}\label{sec:conclusion}
We presented a novel algorithm for predicting the trajectories of road agents in dense and heterogeneous traffic. Our approach is end-to-end, dealing with traffic videos without assuming lane-based driving. Furthermore, we are able to model the interactions between heterogeneous road agents corresponding to cars, buses, pedestrians, two-wheelers, three-wheelers, and animals. We use an LSTM-CNN hybrid network to model two kinds of weighted interactions between road agents: horizon-based and heterogeneous-based. We demonstrate the benefits of our model over state-of-the-art trajectory prediction methods on standard datasets and on a novel dense traffic dataset. We observe up to $30\%$ improvement in prediction accuracy.

Our work has some limitations. Our model design is motivated by some of the characteristics observed in dense heterogeneous traffic. As a result, we do not outperform prior methods on sparse or homogeneous traffic videos, although our prediction results are comparable to prior methods. In addition, modeling heterogeneous constraints requires the knowledge of the shapes and sizes of different road agents. This information could be tedious to collect. In the future, we plan to design a system that eliminates the need for ground truth trajectory data and can directly predict the trajectories from an input video. We also intend to use TraPHic for autonomous navigation in dense traffic.


\section{Acknowledgments}
\vspace{-5pt}
This research is supported in part by ARO grant W911NF19-1- 0069, Alibaba Innovative Research (AIR) program, and Intel.

{\small
\bibliographystyle{ieee}
\bibliography{refs}
}

\end{document}